\documentclass[]{article}
\usepackage[letterpaper]{geometry}
\usepackage{mtsummit2021}
\usepackage{times}
\usepackage{url}
\usepackage{latexsym}
\usepackage{natbib}
\usepackage{layout}
\usepackage{graphicx}
\usepackage{multirow}
\usepackage{threeparttable}
\usepackage{booktabs}
\usepackage{makecell}
\usepackage[utf8]{inputenc}
\usepackage[main=english,russian,arabic]{babel}
\usepackage{xcolor}

\definecolor{green}{RGB}{0,150,0}
\definecolor{red}{RGB}{220,0,0}

\begin{document}

% \amtaHeader{x}{x}{xxx-xxx}{2015}{45-character paper description goes here}{Author(s) initials and last name go here}
\title{\bf Multi-Domain Adaptation in Neural Machine Translation Through Multidimensional Tagging}

\author{\name{\bf Emmanouil Stergiadis} \hfill  \addr{emmanouil.stergiadis@booking.com}\\
        \name{\bf Satendra Kumar} \hfill \addr{satendra.kumar@booking.com}\\
        \name{\bf Fedor Kovalev} \hfill \addr{fedor.kovalev@booking.com}\\
        \name{\bf Pavel Levin} \hfill \addr{pavel.levin@booking.com}\\

}

\maketitle
\pagestyle{empty}

\begin{abstract}
While NMT has achieved remarkable results in the last 5 years, production systems come with strict quality requirements in arbitrarily niche domains that are not always adequately covered by readily available parallel corpora. This is typically addressed by training domain specific models, using fine-tuning methods and some variation of back-translation on top of in-domain monolingual corpora. However, industrial practitioners can rarely afford to focus on a single domain. A far more typical scenario includes a set of closely related, yet succinctly different sub-domains. At Booking.com, we need to translate property descriptions, user reviews, as well as messages, (for example those sent between a customer and an agent or property manager). An editor might need to translate articles across a set of different topics. An e-commerce platform would typically need to translate both the description of each item and the user generated content related to them. To this end, we propose MDT: a novel method to simultaneously fine-tune on several sub-domains by passing multidimensional sentence-level information to the model during training and inference. We show that MDT achieves results competitive to N specialist models each fine-tuned on a single constituent domain, while effectively serving all N sub-domains, therefore cutting development and maintenance costs by the same factor. Besides BLEU (industry standard automatic evaluation metric known to only weakly correlate with human judgement) we also report rigorous human evaluation results for all models and sub-domains as well as specific examples that better contextualise the performance of each model in terms of adequacy and fluency. To facilitate further research, we plan to make the code available upon acceptance.
\end{abstract}

\section{Introduction}

Neural machine translation (NMT) has achieved remarkable results in recent years. A strong testament to its success and efficacy is the increasingly widespread industrial adoption of NMT solutions \cite{johnson2017google, levin2017toward, crego2016systran}. Model parameter estimation in NMT architectures \citep{bahdanau2015neural, gehring2017convolutional, vaswani2017attention} is still largely a supervised learning problem which requires large amounts of translated sentence pairs (parallel data). Obviously, acquiring a sufficient number of high quality parallel sentences in order to train a functional domain-specific NMT system can be prohibitively expensive; especially, if one needs to develop such systems for several domains across different language pairs. On the other hand, large quantities of untranslated in-domain content (monolingual data) are often readily available. 

Various domain adaptation strategies have been developed to address the low-resource setting of niche domains \citep{chu-wang-2018-survey}. Some of the more popular approaches involve generating synthetic in-domain data with the help of existing monolingual corpora, and using that data to fine-tune the more general NMT systems \cite{sennrich2016improving}. 

\begin{figure}[t]
    \centering
    \includegraphics[width=0.8\linewidth]{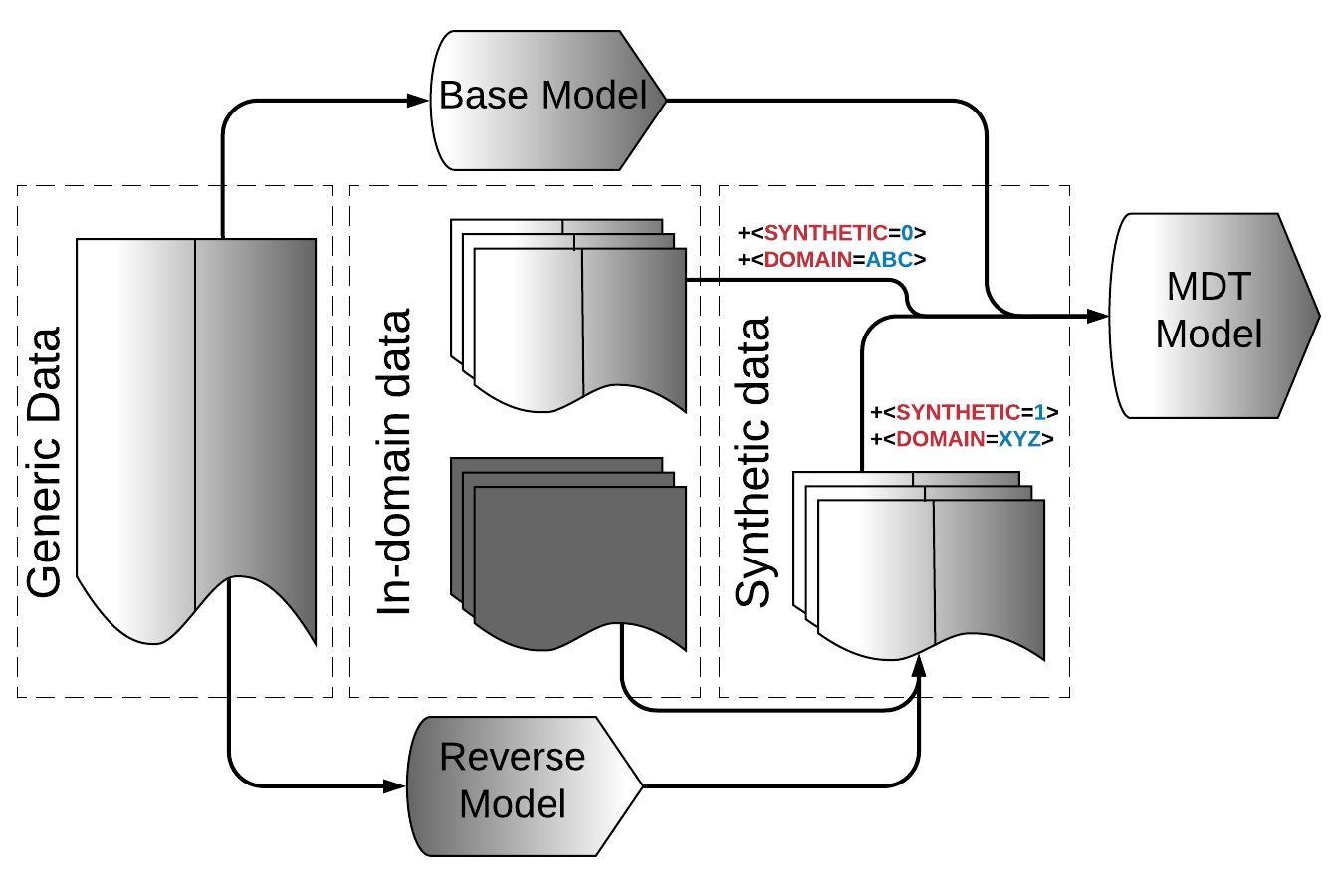}
    \caption{Schematic diagram of MDT in our setting. We use generic parallel data to train a base source-target and a reverse target-source models. We then back-translate target language monolingual in-domain data using the reverse model, and mix it with upsampled in-domain parallel data to fine-tune the base model. The data is tagged with two special tokens: $<$SYNTHETIC=\{0,1\}$>$, and $<$DOMAIN=\{reviews,messaging,descriptions\}$>$.}
    \label{fig:mdt}
\end{figure}

In real-world scenarios practitioners often need to deploy translation engines for several closely related, yet different sub-domains. For example, an online travel marketplace needs to translate offering descriptions, user-generated reviews and customer service communications, all related to travel, but all having different linguistic nuances. This fragmentation is further compounded by the company's need to provide services across many distinct languages. It can be very expensive or outright impossible to develop and maintain separate translation pipelines for every combination of language and sub-domain.

We propose a new method for training models which are simultaneously fine-tuned on several closely related, yet succinctly different sub-domains. We show that those models achieve competitive (and often superior) results to single domain fine-tuned baselines while effectively serving $N$ use cases, therefore cutting development and maintenance costs by a factor of $N$.

\section{Related Work}\label{sec:rel_work}
Our work builds on a growing body of domain adaptation research, mainly related to fine-tuning through tagged back-translation. %Domain fine-tuning is the process of adapting a model which was trained on out-of-domain parallel data to be suitable to specific new or underrepresented domain(s).% Similar to domain fine-tuning, an analogous process has been successfully used to initialize low-resource language pair models \cite{zoph2016transfer}.

\subsection{Domain tagging}
There are a number of research directions related to using tags (or special tokens) within NMT, primarily as a way to pass additional information to the model. Practically speaking, these are attractive approaches as they usually do not require any special modifications to off-the-shelf translation software. The majority of use cases tag sentences on the \textit{source} side: \citet{kobus-etal-2017-domain} use them to control domain, \citet{sennrich-etal-2016-controlling} the politeness, \citet{yamagishi-etal-2016-controlling} the voice and \citet{elaraby2018gender} the gender of translations. The idea also features in multilingual NMT models, for example \citet{johnson2017google} tag training examples according to which translation pair they belong to. An alternative approach by \citet{britz2017effective} prepends the domain tag to the training input on the \textit{target} side, thus forcing the decoder to predict the domain based on the source sentence alone. 

\subsection{Back-Translation for Domain Adaptation}
Back-translation (BT) is a form of semi-supervised learning that can be used to fine-tune both statistical \cite{bertoldi2009domain, bojar2011improving} and neural \citep{sennrich2016improving} machine translation models to new domains. The idea behind this technique is to augment limited parallel in-domain data with a synthetic corpus produced by translating  monolingual data from the \textit{target} language using a \textit{target-to-source} translation system. A synthetic corpus produced via back-translation will have machine-generated source sentences ``translated to" human-written in-domain targets. BT model fine-tuning then becomes a three-stage process: first, genuine parallel data is used to train a reverse model in the target-to-source direction; second, that reverse model is used to translate target-side in-domain monolingual data into the source language; third, synthetic data is used in combination with few truly parallel in-domain samples to fine-tune the base source-to-target model. This simple approach works surprisingly well in practice \cite{bojar-etal-2018-findings, barrault-etal-2019-findings}.

Recent research showed that the details of how we generate the synthetic BT data matter a lot \citep{edunov2018understanding, imamura2018enhancement}. Specifically, the authors find that randomized sampling and noising is preferable to plain beam search. \citet{edunov2018understanding} hypothesise that the improvement is due to randomization contributing to the source-side diversification of the synthetic data. \citet{caswell2019tagged}, on the other hand, suggest that synthetic data adds both helpful and harmful signals, which sampling and noising BT strategies help the model to separate. The TaggedBT technique which they introduce achieves competitive results by simply tagging synthetic data with a special token indicating that the data is machine-generated.

\begin{table}
\centering
%\resizebox{\linewidth}{!}{
{\begin{tabular}{rccc}
\toprule
\multicolumn{1}{l}{}& \textbf{Arabic}& \textbf{German}& \textbf{Russian} \\
\toprule
\multicolumn{1}{c}{\textbf{Parallel}}&&&\\
%\midrule
Generic & 71M & 92M & 87M \\
Reviews & 98k & 63k & 136k \\
Messaging & 73k & 76k & 87k \\
Descriptions & 60k & 72k & 80k \\
\midrule
\multicolumn{1}{c}{\textbf{Monolingual}} &&&\\
%\midrule
Reviews & 1M & 1M & 1M \\
Messaging & 1M & 1M & 1M \\
Descriptions & 1M & 1M & 1M \\
\bottomrule
\end{tabular}
}

\caption{Parallel and monolingual sentences used in our experiments.}
\label{tab:data}
\end{table}

\begin{table*}[]
\centering
\resizebox{\textwidth}{!}{
\begin{tabular}{rccc|ccc|ccc||ccc}
\multicolumn{1}{l}{} & \multicolumn{3}{c|}{\textbf{Reviews}}& \multicolumn{3}{c|}{\textbf{Messaging}}& \multicolumn{3}{c||}{\textbf{Descriptions}}& \multicolumn{3}{c}{\textbf{Average}}\\
\multicolumn{1}{l}{} & \textbf{AR} & \textbf{DE} & \textbf{RU} & \textbf{AR} & \textbf{DE} & \textbf{RU} & \textbf{AR} & \textbf{DE} & \textbf{RU} & \textbf{AR} & \textbf{DE} & \textbf{RU}\\
\multicolumn{1}{l}{\large \textbf{Human score}}& \multicolumn{3}{c|}{\textbf{}}& \multicolumn{3}{c|}{\textbf{}}& \multicolumn{3}{c||}{\textbf{}}\\
\toprule
Base model & 3.65& 3.73& 3.50& 3.27& 3.44& 3.18& 2.67& 3.28& 2.95&3.20&3.48&3.21\\ 
\midrule
+top10   & \makecell{\textbf{3.75}\\\textbf{(+.10)}}&
\makecell{3.80\\(+.07)}&
\makecell{3.57\\(+.07)}&
\makecell{3.36\\(+.09)}&
\makecell{3.65\\(+.19)}&
\makecell{3.53\\(+.35)}&
\makecell{3.02\\(+.35)}&
\makecell{3.70\\(+.42)}&
\makecell{2.95\\(+.00)}&
\makecell{3.38\\(+.18)}&
\makecell{3.71\\(+.23)}& \makecell{3.47\\(+.14)}\\[12pt]
+MDT & \makecell{3.72\\(+.07)}& \makecell{\textbf{3.88}\\\textbf{(+.15)}}&
\makecell{\textbf{3.62}\\\textbf{(+.12)}}&
\makecell{\textbf{3.49}\\\textbf{(+.22)}}&
\makecell{\textbf{3.78}\\\textbf{(+.34)}}&
\makecell{\textbf{3.53}\\ \textbf{(+.35)}}&
\makecell{\textbf{3.20}\\ \textbf{(+.53)}}&
\makecell{\textbf{3.73}\\ \textbf{(+.45)}}&
\makecell{\textbf{3.04}\\ \textbf{(+.09)}}&
\makecell{\textbf{3.47}\\ \textbf{(+.27)}}&
\makecell{\textbf{3.80}\\ \textbf{(+.31)}}&
\makecell{\textbf{3.40}\\ \textbf{(+.19)}}\\
%\multicolumn{10}{c}{}\\
\multicolumn{1}{l}{\large \textbf{BLEU score}} & \multicolumn{9}{r}{}\\
\toprule
Base model & 42.95& 43.63& 38.25& 39.01& 44.18& 41.18& 45.00& 45.97& 38.92&42.32&44.60&39.45\\
\midrule
+top10&
\makecell{\textbf{42.95}\\\textbf{(+0.00)}}&
\makecell{44.99\\(+1.36)}&
\makecell{38.35\\(+0.10)}&
\makecell{41.93\\(+2.92)}&
\makecell{\textbf{50.19}\\\textbf{(+6.01)}}&
\makecell{41.15\\(-0.03)}&
\makecell{45.35\\(+0.35)}&
\makecell{\textbf{50.98}\\\textbf{(+5.01)}}&
\makecell{37.84\\(-1.08)}&
\makecell{43.41\\(+1.09)}&
\makecell{48.72\\(+4.13)}&
\makecell{39.11\\(-0.34)}\\[12pt]
+MDT &
\makecell{42.61\\-0.34}&
\makecell{\textbf{46.34}\\\textbf{(+2.71)}}&
\makecell{\textbf{41.12}\\\textbf{(+2.87)}}&
\makecell{\textbf{47.09}\\\textbf{(+8.08)}}&
\makecell{49.85\\(+5.67)}&
\makecell{\textbf{43.19}\\\textbf{(+2.01)}}&
\makecell{\textbf{46.54}\\\textbf{(+1.54)}}&
\makecell{50.84\\(+4.87)}&
\makecell{\textbf{39.14}\\\textbf{(+0.22)}}&
\makecell{\textbf{45.41}\\\textbf{(+3.09)}}&
\makecell{\textbf{49.01}\\\textbf{(+4.41)}}&
\makecell{\textbf{41.15}\\\textbf{(+1.70)}}\\
\end{tabular}}
\caption{Human evaluations and BLEU scores for the multi-domain adaptation experiments. MDT (our method) is competitive (and on average superior) against the strong fine-tuning baseline (\textit{top10} from \citep{edunov2018understanding}) despite having significantly lower training and deployment costs.}\label{tab:results}
\end{table*}

\section{Multidimensional tagging}\label{sec:mdt}
As discussed in Section \ref{sec:rel_work}, introducing special tokens in the training data has been independently useful at passing content-specific information (e.g.\ domain, voice, gender, etc.) and data-specific information (e.g.\ whether a given data point is synthetic). The current work extends this idea into the  multidimensional setting. Whenever several meaningful dimensions describing the data are available at inference and training time, we can encode that information with special tokens indicating the values along each of the dimensions (Figure \ref{fig:mdt}).

A real-world multi-domain adaptation setting lends itself very naturally to the MDT approach. For example, domain or topic is one such dimension, whether or not the data is synthetic is another. The definition of a synthetic sample may also differ between applications. Back-translation as used in this work is an obvious way of generating such samples, but so can be pseudo-alignment \citep{imankulova2017improving,schwenk2019ccmatrix}. A hybrid dataset may include samples from all three origins (genuine, machine translated and pseudo-aligned) and a tag can help the model differentiate between them. Lastly, multilingual models where the source languages are not trivially different, can be boosted with a language tag\footnote{Independent experiments (not shown in this work) have shown improved results when a Portuguese model is enhanced with a tag denoting a Brazilian versus a European Portuguese author.}. It is therefore clear that although our experiments only cover a two-dimensional setting with the attributes mentioned above (data domain and source), multidimensional tagging can be extended to cover other data aspects.

\section{Experimental Setup}
This section describes our data sources, model architecture, and synthetic data generation and mixing strategies that we employ in our experiments. Our principal goal is to evaluate MDT fine-tuning approach as a scalable alternative to state-of-the-art domain fine-tuning for NMT.

\subsection{Data}
We run our experiments on three language pairs (Arabic-English, German-English and Russian-English) which span three different scripts. Our parallel data sources include a large generic corpus which is a mixture of publicly available and in-house data\footnote{The publicly available portion of our data was sourced from \url{http://opus.nlpl.eu/} \cite{tiedemann2012parallel}}, as well as three much smaller domain-specific parallel datasets (Table \ref{tab:data}). The monolingual data which we use to create back-translated models contains 1M proprietary text segments for each language and domain. All three domains (``Reviews", ``Messaging" and ``Descriptions") are travel-related, and in fact could be considered as sub-domains of a more general ``Travel" domain. Nevertheless, they all exhibit distinct linguistic characteristics which makes it challenging to treat them as a single domain. Appendix \ref{sec:app_examples} provides examples of sentences from different data sources.

\subsection{Synthetic data generation}
We generate all synthetic data using a target-source reverse model trained purely on the generic parallel corpus. According to prior experiments we found \textit{top10}\footnote{Our fine-tuned \textit{top10} baseline was actually our customer-facing production system at the time for several languages.} method from \citet{edunov2018understanding} to be the best-performing domain adaptation method, and we use it as the main approach to benchmark against. Because we do have limited in-domain parallel data, our fine-tuning parallel data is not purely synthetic, but a mix of synthetic and genuine (which we upsample to reach 1:1 composition).

\paragraph{top10} Following \citet{edunov2018understanding} we use our reverse target-source models to translate monolingual data back to English, but at the generation stage we \textit{sample} from the next token distribution instead of using beam search to approximate MAP translation. At each sampling step we only consider top 10 most probable candidates.

\paragraph{MDT} As described in Section \ref{sec:mdt}, we extend the idea of tagged BT \cite{caswell2019tagged} to multi-attribute setting by prepending source-side tags which qualify various aspects of the data. Specifically, in this experiment we tag the data according to two characteristics: (1) whether it is synthetically generated or genuine, (2) which sub-domain it belongs to. Both types of tags are treated just like any other tokens, i.e.\ their learned embeddings are stored in the shared source-side embeddings table.

\subsection{Model architecture}
Prior to feeding parallel data into the sequence-to-sequence models, all text is preprocessed using the byte-pair encoding (BPE) tokenization scheme \citep{sennrich2016neural}. Our models follow the transformer-base architecture from \citet{vaswani2017attention} as implemented in OpenNMT-tf\footnote{\url{https://github.com/OpenNMT/OpenNMT-tf}} v1.25 \citep{klein-etal-2017-opennmt} with early stopping based on development sets of 5000 sentences per each use case.

\subsection{Evaluation}\label{sec:eval}
The context of this work is a real-world industrial setting which involves translating large volumes of customer-facing text. Therefore our main evaluation criteria are human-based assessments. The human evaluation was performed by professional translators on a 4-point adequacy Likert scale using 250 samples per language, per domain. Appendix \ref{sec:app_human} provides details of the scoring guidelines that human evaluators follow. Additionally we report case-sensitive BLEU score \citep{papineni-etal-2002-bleu} as implemented by sacreBLEU\footnote{\url{https://github.com/mjpost/sacreBLEU}} \cite{post-2018-call}.

\section{Results}

\subsection{Multi-domain adaptation}
Table \ref{tab:results} summarizes our multi-domain adaptation results. On average MDT does not only match, but in fact outperforms the strong \textit{top10} \citep{edunov2018understanding} baseline. As mentioned in Section \ref{sec:eval}, given the production quality requirement of our systems we consider human scoring the gold standard for evaluating translations, not the BLEU score alone. Most human and BLEU scores do rank-wise agree, but there are some exceptions. Specifically the German-English MDT model does better than the respective \textit{top10} models on Messaging and Descriptions domains according to the human evaluators, however it is not reflected in the BLEU scores.

\subsection{Ablation experiment}
In order to assess the role of tags, we perform an ablation experiment for German language, in which we compare the MDT performance to that of a model trained without the tags (but on the same mix of training data). It appears that the tags indeed on average improve the performance (Table \ref{tab:ablation}). The models without tags perform worse on ``Reviews" and ``Messaging" domains according to human evaluations, and on all three domains according to the BLEU score evaluations. 
%It seems that human evaluators disagree with the BLEU score for one specific sub-topic (``Descriptions").

\begin{table}[h!]
    \centering
    %\resizebox{0.8\linewidth}{!}{
{
\begin{tabular}{rll}
    \toprule
    &\textbf{Human score} & \textbf{BLEU score} \\
    \midrule
    \multicolumn{3}{l}{\textbf{Reviews}} \\
    MDT Model& 3.88 & 46.34\\
    (-tags) & 3.82 (-.06) & 44.24 (-2.10)\\
    \hline
    \multicolumn{3}{l}{\textbf{Messaging}} \\
    MDT Model & 3.78 & 49.85\\
    (-tags) & 3.48 (-.30) & 49.21 (-0.64)\\
    \hline
    \multicolumn{1}{l}{\textbf{Descriptions}} &  & \\
    MDT Model& 3.73 & 50.84\\
    (-tags) & 3.80 (+.07) & 49.79 (-1.05)\\
    \midrule
    \textbf{Average} &
    \multicolumn{1}{c}{-.10} &
    \multicolumn{1}{c}{-1.26}\\
    \bottomrule
\end{tabular}
}
    \caption{The effect of tags removal on human and BLEU score in German-English MDT model.}
    \label{tab:ablation}
\end{table}

\section{Conclusions}
In this work we introduce multidimensional tagging and demonstrate that it can be a scalable solution for multi-domain adaptation in a realistic resource-constrained setting. Somewhat surprisingly we find that MDT models in fact outperform on average our best alternative fine-tuning technique (\textit{top10} from \citet{edunov2018understanding}), even though the alternative method trains a custom model for each sub-topic. Although the present work offers limited empirical evaluations of MDT (two dimensions: 3 sub-domains and 2 data sources; three language pairs), we think that the technique can prove useful in a broader setting. We believe it to be particularly well suited to many real-world scenarios in which practitioners develop solutions for multiple related domains, while leveraging data from different sources, both genuine and synthetic. All experimental results reported in this work follow rigorous human evaluations in addition to the standard BLEU scores assessments.

\bibliographystyle{apalike}
\bibliography{mtsummit2021, anthology}

\begin{thebibliography}{}

\bibitem[{Bahdanau} et~al., 2015]{bahdanau2015neural}
{Bahdanau}, D., {Cho}, K., and {Bengio}, Y. (2015).
\newblock Neural machine translation by jointly learning to align and
  translate.
\newblock In {\em ICLR 2015 : International Conference on Learning
  Representations 2015}.

\bibitem[Barrault et~al., 2019]{barrault-etal-2019-findings}
Barrault, L., Bojar, O., Costa-juss{\`a}, M.~R., Federmann, C., Fishel, M.,
  Graham, Y., Haddow, B., Huck, M., Koehn, P., Malmasi, S., Monz, C.,
  M{\"u}ller, M., Pal, S., Post, M., and Zampieri, M. (2019).
\newblock Findings of the 2019 conference on machine translation ({WMT}19).
\newblock In {\em Proceedings of the Fourth Conference on Machine Translation
  (Volume 2: Shared Task Papers, Day 1)}, pages 1--61, Florence, Italy.
  Association for Computational Linguistics.

\bibitem[Bertoldi and Federico, 2009]{bertoldi2009domain}
Bertoldi, N. and Federico, M. (2009).
\newblock Domain adaptation for statistical machine translation with
  monolingual resources.
\newblock In {\em Proceedings of the fourth workshop on statistical machine
  translation}, pages 182--189.

\bibitem[Bojar et~al., 2018]{bojar-etal-2018-findings}
Bojar, O., Federmann, C., Fishel, M., Graham, Y., Haddow, B., Koehn, P., and
  Monz, C. (2018).
\newblock Findings of the 2018 conference on machine translation ({WMT}18).
\newblock In {\em Proceedings of the Third Conference on Machine Translation:
  Shared Task Papers}, pages 272--303, Belgium, Brussels. Association for
  Computational Linguistics.

\bibitem[Bojar and Tamchyna, 2011]{bojar2011improving}
Bojar, O. and Tamchyna, A. (2011).
\newblock Improving translation model by monolingual data.
\newblock In {\em Proceedings of the Sixth Workshop on Statistical Machine
  Translation}, pages 330--336. Association for Computational Linguistics.

\bibitem[Britz et~al., 2017]{britz2017effective}
Britz, D., Le, Q., and Pryzant, R. (2017).
\newblock Effective domain mixing for neural machine translation.
\newblock In {\em Proceedings of the Second Conference on Machine Translation},
  pages 118--126.

\bibitem[Callison-Burch et~al., 2007]{callison2007meta}
Callison-Burch, C., Fordyce, C., Koehn, P., Monz, C., and Schroeder, J. (2007).
\newblock (meta-) evaluation of machine translation.
\newblock In {\em Proceedings of the Second Workshop on Statistical Machine
  Translation}, pages 136--158. Association for Computational Linguistics.

\bibitem[{Caswell} et~al., 2019]{caswell2019tagged}
{Caswell}, I., {Chelba}, C., and {Grangier}, D. (2019).
\newblock Tagged back-translation.
\newblock In {\em Proceedings of the Fourth Conference on Machine Translation
  (Volume 1: Research Papers)}, pages 53--63.

\bibitem[Chu and Wang, 2018]{chu-wang-2018-survey}
Chu, C. and Wang, R. (2018).
\newblock A survey of domain adaptation for neural machine translation.
\newblock In {\em Proceedings of the 27th International Conference on
  Computational Linguistics}, pages 1304--1319, Santa Fe, New Mexico, USA.
  Association for Computational Linguistics.

\bibitem[Crego et~al., 2016]{crego2016systran}
Crego, J., Kim, J., Klein, G., Rebollo, A., Yang, K., Senellart, J., Akhanov,
  E., Brunelle, P., Coquard, A., Deng, Y., et~al. (2016).
\newblock Systran's pure neural machine translation systems.
\newblock {\em arXiv preprint arXiv:1610.05540}.

\bibitem[{Edunov} et~al., 2018]{edunov2018understanding}
{Edunov}, S., {Ott}, M., {Auli}, M., and {Grangier}, D. (2018).
\newblock Understanding back-translation at scale.
\newblock In {\em EMNLP 2018: 2018 Conference on Empirical Methods in Natural
  Language Processing}, pages 489--500.

\bibitem[Elaraby et~al., 2018]{elaraby2018gender}
Elaraby, M., Tawfik, A.~Y., Khaled, M., Hassan, H., and Osama, A. (2018).
\newblock Gender aware spoken language translation applied to english-arabic.
\newblock In {\em 2018 2nd International Conference on Natural Language and
  Speech Processing (ICNLSP)}, pages 1--6. IEEE.

\bibitem[Gehring et~al., 2017]{gehring2017convolutional}
Gehring, J., Auli, M., Grangier, D., Yarats, D., and Dauphin, Y.~N. (2017).
\newblock Convolutional sequence to sequence learning.
\newblock In {\em Proceedings of the 34th International Conference on Machine
  Learning-Volume 70}, pages 1243--1252. JMLR. org.

\bibitem[Imamura et~al., 2018]{imamura2018enhancement}
Imamura, K., Fujita, A., and Sumita, E. (2018).
\newblock Enhancement of encoder and attention using target monolingual corpora
  in neural machine translation.
\newblock In {\em Proceedings of the 2nd Workshop on Neural Machine Translation
  and Generation}, pages 55--63.

\bibitem[Imankulova et~al., 2017]{imankulova2017improving}
Imankulova, A., Sato, T., and Komachi, M. (2017).
\newblock Improving low-resource neural machine translation with filtered
  pseudo-parallel corpus.
\newblock In {\em Proceedings of the 4th Workshop on Asian Translation
  (WAT2017)}, pages 70--78.

\bibitem[Johnson et~al., 2017]{johnson2017google}
Johnson, M., Schuster, M., Le, Q.~V., Krikun, M., Wu, Y., Chen, Z., Thorat, N.,
  Vi{\'e}gas, F., Wattenberg, M., Corrado, G., et~al. (2017).
\newblock Google’s multilingual neural machine translation system: Enabling
  zero-shot translation.
\newblock {\em Transactions of the Association for Computational Linguistics},
  5:339--351.

\bibitem[{Kingma} and {Ba}, 2014]{kingma2014method}
{Kingma}, D.~P. and {Ba}, J. (2014).
\newblock Adam: A method for stochastic optimization.
\newblock In {\em Proceedings of the 3rd International Conference on Learning
  Representations (ICLR)}.

\bibitem[Klein et~al., 2017]{klein-etal-2017-opennmt}
Klein, G., Kim, Y., Deng, Y., Senellart, J., and Rush, A. (2017).
\newblock {O}pen{NMT}: Open-source toolkit for neural machine translation.
\newblock In {\em Proceedings of {ACL} 2017, System Demonstrations}, pages
  67--72, Vancouver, Canada. Association for Computational Linguistics.

\bibitem[Kobus et~al., 2017]{kobus-etal-2017-domain}
Kobus, C., Crego, J., and Senellart, J. (2017).
\newblock Domain control for neural machine translation.
\newblock In {\em Proceedings of the International Conference Recent Advances
  in Natural Language Processing, {RANLP} 2017}, pages 372--378, Varna,
  Bulgaria. INCOMA Ltd.

\bibitem[{Levin} et~al., 2017a]{levin2017toward}
{Levin}, P., {Dhanuka}, N., {Khalil}, T., {Kovalev}, F., and {Khalilov}, M.
  (2017a).
\newblock Toward a full-scale neural machine translation in production: the
  booking.com use case.
\newblock In {\em Proceedings of MT Summit XVI}, volume~2, pages 39--49.

\bibitem[{Levin} et~al., 2017b]{levin2017machine}
{Levin}, P., {Dhanuka}, N., and {Khalilov}, M. (2017b).
\newblock Machine translation at booking. com: Journey and lessons learned.
\newblock In {\em Proceedings of the 20th International Conference of the
  European Association for Machine Translation (EAMT)}.

\bibitem[Papineni et~al., 2002]{papineni-etal-2002-bleu}
Papineni, K., Roukos, S., Ward, T., and Zhu, W.-J. (2002).
\newblock {B}leu: a method for automatic evaluation of machine translation.
\newblock In {\em Proceedings of the 40th Annual Meeting of the Association for
  Computational Linguistics}, pages 311--318, Philadelphia, Pennsylvania, USA.
  Association for Computational Linguistics.

\bibitem[Post, 2018]{post-2018-call}
Post, M. (2018).
\newblock A call for clarity in reporting {BLEU} scores.
\newblock In {\em Proceedings of the Third Conference on Machine Translation:
  Research Papers}, pages 186--191, Belgium, Brussels. Association for
  Computational Linguistics.

\bibitem[Schwenk et~al., 2019]{schwenk2019ccmatrix}
Schwenk, H., Wenzek, G., Edunov, S., Grave, E., and Joulin, A. (2019).
\newblock Ccmatrix: Mining billions of high-quality parallel sentences on the
  web.
\newblock {\em arXiv preprint arXiv:1911.04944}.

\bibitem[Sennrich et~al., 2016]{sennrich-etal-2016-controlling}
Sennrich, R., Haddow, B., and Birch, A. (2016).
\newblock Controlling politeness in neural machine translation via side
  constraints.
\newblock In {\em Proceedings of the 2016 Conference of the North {A}merican
  Chapter of the Association for Computational Linguistics: Human Language
  Technologies}, pages 35--40, San Diego, California. Association for
  Computational Linguistics.

\bibitem[{Sennrich} et~al., 2016a]{sennrich2016improving}
{Sennrich}, R., {Haddow}, B., and {Birch}, A. (2016a).
\newblock Improving neural machine translation models with monolingual data.
\newblock In {\em Proceedings of the 54th Annual Meeting of the Association for
  Computational Linguistics (Volume 1: Long Papers)}, volume~1, pages 86--96.

\bibitem[{Sennrich} et~al., 2016b]{sennrich2016neural}
{Sennrich}, R., {Haddow}, B., and {Birch}, A. (2016b).
\newblock Neural machine translation of rare words with subword units.
\newblock In {\em Proceedings of the 54th Annual Meeting of the Association for
  Computational Linguistics (Volume 1: Long Papers)}, volume~1, pages
  1715--1725.

\bibitem[Tiedemann, 2012]{tiedemann2012parallel}
Tiedemann, J. (2012).
\newblock Parallel data, tools and interfaces in opus.
\newblock In {\em Lrec}, volume 2012, pages 2214--2218.

\bibitem[{Vaswani} et~al., 2017]{vaswani2017attention}
{Vaswani}, A., {Shazeer}, N., {Parmar}, N., {Uszkoreit}, J., {Jones}, L.,
  {Gomez}, A.~N., {Kaiser}, L., and {Polosukhin}, I. (2017).
\newblock Attention is all you need.
\newblock In {\em Proceedings of the 31st International Conference on Neural
  Information Processing Systems}, pages 5998--6008.

\bibitem[White et~al., 1994]{white1994arpa}
White, J., O’Connell, T., and O’Mara, F. (1994).
\newblock The arpa mt evaluation methodologies: evolution, lessons, and future
  approaches.
\newblock In {\em Proceedings of the First Conference of the Association for
  Machine Translation in the Americas}, pages 193--205.

\bibitem[Yamagishi et~al., 2016]{yamagishi-etal-2016-controlling}
Yamagishi, H., Kanouchi, S., Sato, T., and Komachi, M. (2016).
\newblock Controlling the voice of a sentence in {J}apanese-to-{E}nglish neural
  machine translation.
\newblock In {\em Proceedings of the 3rd Workshop on {A}sian Translation
  ({WAT}2016)}, pages 203--210, Osaka, Japan. The COLING 2016 Organizing
  Committee.

\end{thebibliography}

%\clearpage
\appendix
\section*{Supplementary Material}
\section{Human evaluations criteria}\label{sec:app_human}
\begin{figure*}[b]
  \centering
  \includegraphics[width=\linewidth]{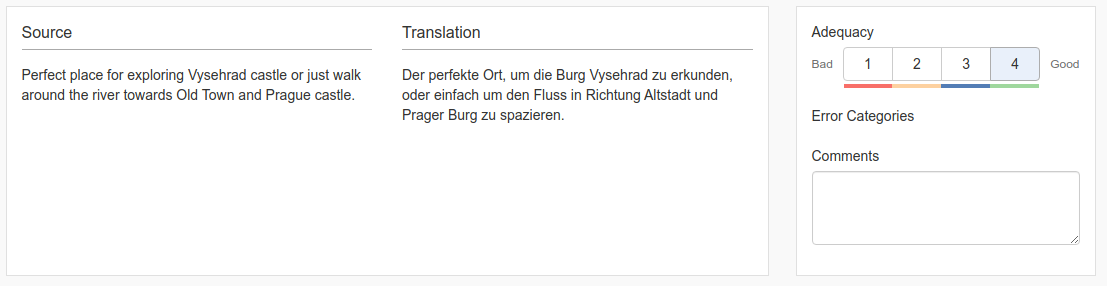}
  \caption{A screenshot of the internal human evaluation tool used by the language specialists.}\label{fig:tool}
\end{figure*}
Each reported human evaluation reading is based on a random test set of 250 text samples which are evaluated by professional translators. Even though all translators were aligned and calibrated during previous evaluations, all sentences from the sample are always sent to the same individual translator to preserve consistency. We use an internally built tool (Figure \ref{fig:tool}) which allows scoring on a four-point Likert scale, a modified version of the "Accuracy" dimension of the Fluency/Adequacy framework \cite{white1994arpa, callison2007meta, levin2017machine}. We observed that fluency is almost never an issue in neural machine translation, so we do not score it explicitly. The following are the scoring guidelines for the four-point accuracy scale that are given to the translators:

\begin{table}[h]
\centering
\resizebox{\linewidth}{!}{
\begin{tabular}{cp{1\linewidth}} % PLAY WITH THIS NUMBER TO FIT TO PAGE
\toprule
\begin{tabular}{c}\textbf{4}\end{tabular}& \begin{tabular}{p{\linewidth}}All aspects of the review are comprehensible.\end{tabular}\\
\midrule
\begin{tabular}{l}\textbf{3}\end{tabular}& \begin{tabular}{p{\linewidth}}The fundamental information provided is accurately conveyed in the translation. Minor errors in non-essential supplementary information that are vague or obscured, but do not contend with the core of the meaning in the description, are allowed.\end{tabular}\\
\midrule
\begin{tabular}{l}\textbf{2}\end{tabular}& \begin{tabular}{p{\linewidth}}The fundamental information provided is obscured/distorted. The translation either indicates different factual information to what is present in the source, or the translation introduces incorrect information.\end{tabular}\\
\midrule
\begin{tabular}{l}\textbf{1}\end{tabular}&\begin{tabular}{p{\linewidth}} The translation does not make any sense, and/or does not even allude to the core of the source text.\end{tabular}\\
\bottomrule
\end{tabular}
}
\end{table}

% \begin{table}[h]
% \centering
% \resizebox{\linewidth}{!}{
% \begin{tabular}{cp{1.24\linewidth}} % PLAY WITH THIS NUMBER TO FIT TO PAGE
% \toprule
% \begin{tabular}{c}\textbf{4}\end{tabular}& \begin{tabular}{p{\linewidth}}All aspects of the review are comprehensible.\end{tabular}\\
% \midrule
% \begin{tabular}{l}\textbf{3}\end{tabular}& \begin{tabular}{p{\linewidth}}The fundamental information provided is accurately conveyed in the translation. Minor errors in non-essential supplementary information that are vague or obscured, but do not contend with the core of the meaning in the description, are allowed.\end{tabular}\\
% \midrule
% \begin{tabular}{l}\textbf{2}\end{tabular}& \begin{tabular}{p{\linewidth}}The fundamental information provided is obscured/distorted. The translation either indicates different factual information to what is present in the source, or the translation introduces incorrect information.\end{tabular}\\
% \midrule
% \begin{tabular}{l}\textbf{1}\end{tabular}&\begin{tabular}{p{\linewidth}} The translation does not make any sense, and/or does not even allude to the core of the source text.\end{tabular}\\
% \bottomrule
% \end{tabular}
% }
% \end{table}

\section{Reproducibility}\label{sec:app_rep}
Prior to feeding parallel data into the sequence-to-sequence models, all text is pre-processed using byte-pair encoding (BPE) tokenization scheme \citep{sennrich2016neural}. For all language pairs the BPE vocabulary size is set to 32k. For EN-DE language pair the vocabulary is learned jointly, while for EN-RU and EN-AR we use separate 32k vocabularies due to different alphabets in source and target. All our models follow the transformer-base architecture as described in  \citet{vaswani2017attention} and implemented in OpenNMT-tf software \citep{klein-etal-2017-opennmt}\footnote{\url{https://github.com/OpenNMT/OpenNMT-tf}}. We trained the models using Adam \cite{kingma2014method} optimizer with $\beta_1=0.9$ and $\beta_2=0.998$ with label smoothing set to 0.1 and noam decay with an initial learning rate of 2.0. While no hyper-parameter tuning is done, early stopping is based on a dev set of 5000 sentences. Furthermore, we use an effective batch size of 25,000 tokens accumulated over different GPUs and keep training until validation loss does not decrease for two consecutive steps. We select the checkpoint with minimum sentence level validation loss - therefore completely ignoring BLEU at model selection. We report both BLEU and human evaluation results using beam width equal to four on a separate test set.

Training our base models took around 5 days using 8 NVIDIA V100 GPUs. Fine-tuning (both the single-domain baseline and the multi-domain MDT variant) took around 16 hours on a single GPU of the same model showing that there is no noticeable difference in training time. Inference time is the same for all models and only depends on sequence length. 

%Although we cannot share our training data for legal and privacy reasons, we attach all pre-processing and training scripts necessary for reproducibility as supplementary material to this paper. It will be made public after the anonymity period.
\section{Text samples}\label{sec:app_examples}
The table below provides a few typical text samples from each domain for each of the three source languages. We also show English reference (human) translation as well as translation outputs from each of the three engines: base model, domain fine-tuned model (top10) and MDT (our method).

\vspace{1cm}
\begin{table*}[h]
    \centering
\resizebox{\textwidth}{!}{
    \begin{tabular}{rl}
    %\toprule
    \multicolumn{2}{l}{\Large\textbf{Reviews}} \\
    \midrule
        \textbf{Source} & \foreignlanguage{russian}{Были всего одну ночь, поэтому в полной мере оценить не смогли.}\\
         \textbf{Reference}&We only stayed there for one night, so we couldn’t fully appreciate it.\\
         \textbf{Base model}&\textcolor{red}{There was only} one night, so we could not fully appreciate it.\\
         \textbf{top10}&\textcolor{green}{We were there only for} one night, so we couldn't fully appreciate it.\\
         \textbf{MDT}&\textcolor{green}{We were only there for} one night, so we could not fully appreciate it.\\
         &\\

        \textbf{Source} & {die Abwesenheit von Personal der Raum lies sich nicht heizen}\\
         \textbf{Reference}&absence of staff the room could not be heated\\
         \textbf{Base model}&the absence of personnel \textcolor{red}{in} the room could not be heated\\
         \textbf{top10}&the absence of staff the room could not be heated\\
         \textbf{MDT}&the absence of staff the room could not be heated\\
         &\\
         
         \textbf{Source} &\AR{مكانه فقط}\\
         \textbf{Reference}&Its location only\\
         \textbf{Base model}&\textcolor{red}{Just his place.}\\
         \textbf{top10}&Its location only\\
         \textbf{MDT}&Its location only\\
\bottomrule
\end{tabular}}
%    \caption{Caption}
    \label{tab:examples_reviews}
\end{table*}

\vspace{1cm}
\begin{table*}[h]
    \centering
\resizebox{\textwidth}{!}{
    \begin{tabular}{rl}
    %\toprule
        \multicolumn{2}{l}{\Large\textbf{Messaging}} \\
    \midrule
        \textbf{Source} &
        \foreignlanguage{russian}{если можно не выше второго этажа спасибо}\\
         \textbf{Reference}&If possible not higher than the second floor thank you.\\
         \textbf{Base model}&\textcolor{red}{If you can't go above the} second floor thank you\\
         \textbf{top10}&\textcolor{green}{if possible not higher than} the second floor thank you\\
         \textbf{MDT}&\textcolor{green}{if possible no higher than} the second floor thank you\\
         &\\
    
        \textbf{Source} &
        wir möchten Elli, unsere Dalmatiner Hündin mitbringen.\\
         \textbf{Reference}&we would like to bring Elli, our Dalmatian dog.\\
         \textbf{Base model}&We'd like to bring Elli our Dalmatian \textcolor{red}{bitch}.\\
         \textbf{top10}&we would like to bring Elli, our Dalmatian \textcolor{green}{dog}.\\
         \textbf{MDT}&we would like to bring our Dalmatian \textcolor{green}{dog} Elli.\\
         &\\
         
        \textbf{Source} &
        \AR{مرحبا هل الدفع بالليرة ؟ وكم التكلفة لثلاث ليالي بالليرة}\\
         \textbf{Reference}&Hello, is the payment in Lira? What is the cost for three nights in Lira?\\
         \textbf{Base model}&Hey. \textcolor{red}{Is it a lira}? How much for three nights a lira?\\
         \textbf{top10}&Hello! Is the payment in \textcolor{red}{pounds}? And how much is it for 3 nights in lira\\
         \textbf{MDT}&Hello Is the payment in lira? And how much it cost for 3 nights \textcolor{red}{per} lira.\\
\bottomrule
\end{tabular}}
%    \caption{Caption}
    \label{tab:examples_messaging}
\end{table*}

%\clearpage
\begin{table*}[h]
    \centering
\resizebox{\textwidth}{!}{
    \begin{tabular}{rp{11.5cm}}
    %\toprule
        \multicolumn{2}{l}{\Large\textbf{Descriptions}} \\
        \midrule
    
        \textbf{Source} & \foreignlanguage{russian}{Просторные апартаменты обставленные в современном стиле, но при этом по домашнему уютные.} \\
         \textbf{Reference}&Spacious apartments are fitted in a modern style, but are still cosy like home.\\
         \textbf{Base model}&Spacious apartment with modern furnishings and \textcolor{red}{homelike} interiors.\\
         \textbf{top10}&Spacious apartments furnished in a modern style, but at the same time homely.\\
         \textbf{MDT}&Spacious apartments furnished in a modern style, but at the same time \textcolor{green}{homely}.\\
         &\\

        \textbf{Source} &Feste und Kulinarik auf höchster Ebene garantieren Abwechslung das ganze Jahr!\\
         \textbf{Reference}&Festivals and culinary delights of the highest standard guarantee variety all year round!\\
         \textbf{Base model}&Festive and \textcolor{red}{culinary cuisine} at the highest level guarantees variety all year round!\\
         \textbf{top10}&Festivals and \textcolor{green}{culinary delights} at the highest level guarantee variety all year round!\\
         \textbf{MDT}&Festivals and \textcolor{green}{culinary delights} at the highest level guarantee variety all year round!\\
         &\\
         
         \textbf{Source} &\AR{مكان رائع لإقامه ممتعه يقع فى قرية بورتو ساوث بيتش بالعين السخنه حيث الجو الممتع والطبيعة الخلابة وحيث يتواصل البحر بالجبل والطبيعة الخلابة
}\\
         \textbf{Reference}&A great place for a pleasant stay located in the village of Porto South Beach in Ain Sokhna, where the atmosphere is enjoyable and picturesque nature, and where the sea meets the mountain and picturesque nature\\
         \textbf{Base model}&A great place for an enjoyable stay, located in the village of Porto South Beach with \textcolor{red}{the hot eye}, where the atmosphere is enjoyable and nature is picturesque and where the sea communicates with the mountain and picturesque nature\\
         \textbf{top10}&A great place to stay, located in the village of Porto South Beach in Ain Sokhna, where the atmosphere is pleasant and the nature is wonderful and where the sea communicates with the mountain and the wonderful nature\\
         \textbf{MDT}&A great place for a pleasant stay located in the village of Porto South Beach in Ain Sokhna, where the atmosphere is pleasant and the nature is picturesque and where the sea communicates with the mountain and the picturesque nature\\
\bottomrule
\end{tabular}}
%    \caption{Caption}
    \label{tab:examples_desciptions}
\end{table*}

\small

\end{document}